\documentclass[10pt,twocolumn,letterpaper]{article}

\usepackage{cvpr}              
\definecolor{cvprblue}{rgb}{0.21,0.49,0.74}
\usepackage[pagebackref,breaklinks,colorlinks,allcolors=cvprblue]{hyperref}

\usepackage[dvipsnames]{xcolor}
\definecolor{col1}{RGB}{232, 161, 148}
\definecolor{col2}{RGB}{148, 187, 232}
\usepackage{colortbl}   
\usepackage{multirow}
\usepackage[ruled,vlined]{algorithm2e}
\usepackage[table]{xcolor}
\usepackage{dsfont}
\def\paperID{****} 
\def\confName{CVPR}
\def\confYear{2026}

\newcommand{\name}{RISE}

\title{\name: Single Static Radar-based Indoor Scene Understanding
}
\author{
Kaichen Zhou$^{1}$   
\and 
Laura Dodds$^{1}$
\and
Sayed Saad Afzal$^{1}$
\and 
Fadel Adib$^{1,2}$    
\and
\\
$^1$Massachusetts Institute of Technology \hspace{2em} 
$^2$Cartesian Systems  
}

\begin{document}
\twocolumn[{
\maketitle
\begin{center}
\includegraphics[width=\linewidth]{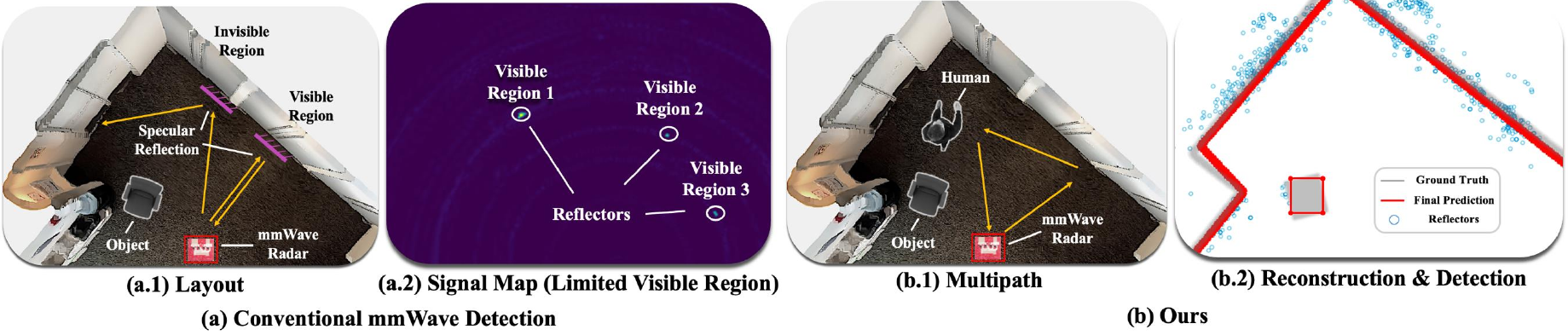}
\end{center}
\vspace{-0.5cm}
\captionsetup{type=figure}
\captionof{figure}{%
\textbf{\name} is the first single-radar system for object-level indoor scene understanding.
Under standard millimeter-wave (mmWave) sensing \textbf{(a.1)}, the radar directly observes only a small portion of the environment, producing sparse visible regions \textbf{(a.2)} due to strong specular reflections.
To overcome this limitation, \name{} performs \textbf{multipath inversion} \textbf{(b.1)} to exploit indirect reflections and recover surfaces that are not directly visible.
Combined with our geometry-aware reconstruction pipeline, \name{} enables accurate and complete indoor layout reconstruction and object detection \textbf{(b.2)}.
}\label{fig:teaser}
\vspace{0.3cm}
}]

\begin{abstract}
Robust and privacy-preserving indoor scene understanding remains a fundamental open problem.
While optical sensors such as RGB and LiDAR offer high spatial fidelity, they suffer from severe occlusions and introduce privacy risks in indoor environments.
In contrast, millimeter-wave (mmWave) radar preserves privacy and penetrates obstacles, but its inherently low spatial resolution makes reliable geometric reasoning difficult.
We introduce \name, the first benchmark and system for single-static-radar indoor scene understanding, jointly targeting layout reconstruction and object detection.
\name{} is built upon the key insight that multipath reflections—traditionally treated as noise—encode rich geometric cues.
To exploit this, we propose a Bi-Angular Multipath Enhancement that explicitly models Angle-of-Arrival and Angle-of-Departure to recover secondary (ghost) reflections and reveal invisible structures.
On top of these enhanced observations, a simulation-to-reality Hierarchical Diffusion framework transforms fragmented radar responses into complete layouts reconstruction and object detection.
Our benchmark contains 50,000 frames collected across 100 real indoor trajectories, forming the first large-scale dataset dedicated to single, static, radar-based indoor scene understanding.
Extensive experiments show that \name{} reduces the Chamfer Distance by 60\% (down to 16 cm) compared to the state of the art in mmWave layout reconstruction, and delivers the first mmWave-based object detection, achieving 58\% IoU.
These results establish \name{} as a new foundation for geometry-aware and privacy-preserving indoor scene understanding using a single static radar. 
Our website and code are available at \url{https://rise-cvpr.github.io}.
\end{abstract}
\vspace{-0.5cm}
    
\section{Introduction}
\label{sec:intro} 

Indoor scene understanding has long been a fundamental research problem in the vision community, with wide-ranging applications in smart homes~\cite{yang2019security}, virtual reality (VR)~\cite{manni2021snap2cad}, and augmented reality (AR)~\cite{yang2013image}. 
Classical approaches to this problem rely on optical sensors such as RGB cameras and LiDAR, which suffer from visual occlusions since they cannot image through common obstructions like walls or objects~\cite{zhang2020object, schneider2010fusing} and also raise privacy concerns in smart home and corporate environments~\cite{lin2016iot}.
To overcome these challenges,  researchers have recently started investigating the use of wireless signals -- like WiFi or millimeter-waves (mmWaves) -- for indoor scene reconstruction since such signals can penetrate everyday occlusions and are much less privacy intrusive than cameras~\cite{cui2023milipoint, fuller2023croma, zhang2024tarss, ding2024radarocc, ho2024rt, ding2024milliflow, bialer2024radsimreal, bang2024radardistill, zhang2023peakconv, li2023azimuth, peng2024transloc4d}. Despite promising early results, existing wireless solutions either suffer from low resolution -- providing only limited patches of the surrounding environment -- or require mounting the wireless sensor on a mobile robot to scan the environment -- which increases deployment overhead, making them less desirable~\cite{
lee2024carb, zhao2024crkd, liu2023echoes, kim2023crn, lin2024rcbevdet, li2024sparse, singh2023depth, wang2023bi, chae2024towards, wu2023nlos, zhang2021building, hernandez2022wifi}.

\begin{figure}[t!]
    \centering
    \includegraphics[width=\linewidth, clip]{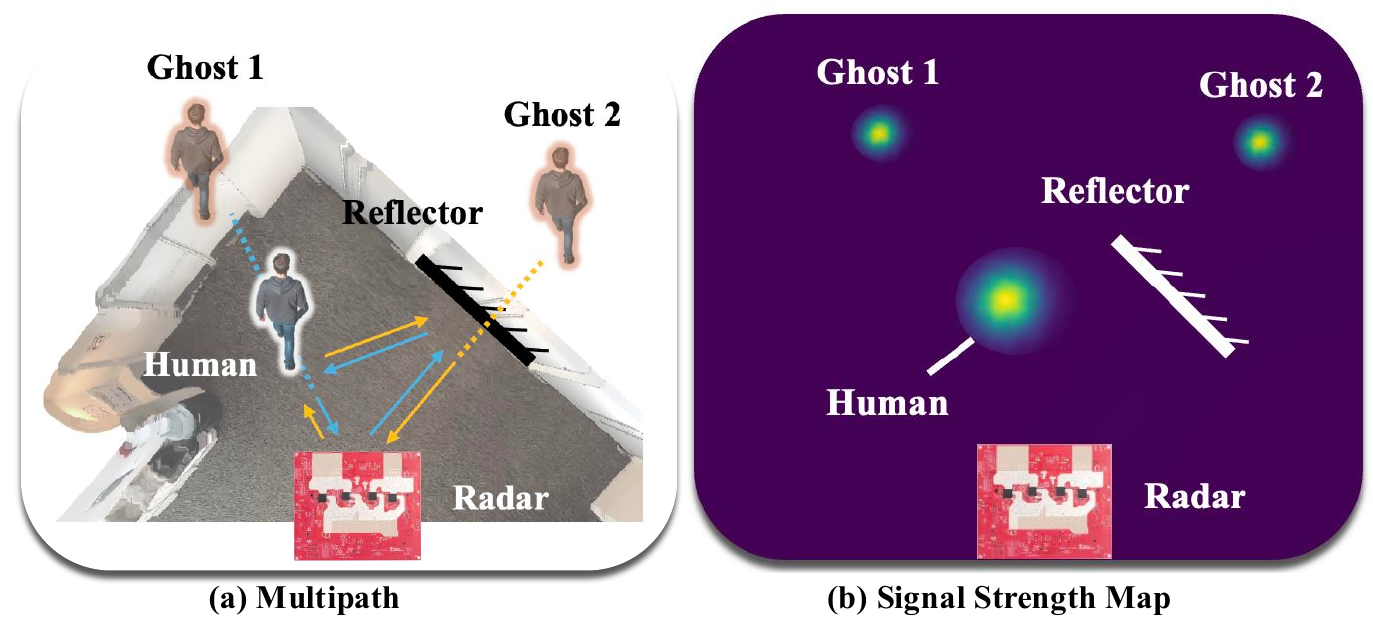} 
    \vspace{-0.6cm}
    \caption{\textbf{Multipath-induced Ghosts:} Figure (a) illustrates the scenario in which ghost points appear. Figure (b) presents the corresponding XY heatmap generated by the radar.}
    \label{fig:teaser2}
    \vspace{-0.5cm}
\end{figure}

In this paper, we ask the following question: \textit{Can we use a single static mmWave wireless device to enable accurate indoor scene understanding?} A positive answer would enable leveraging a typical wireless setup -- such as an existing access point or wireless router already deployed in smart homes -- for this purpose.  
A main challenge to enable indoor layout reconstruction via static mmWave sensing is specularity \cite{qian2022millimirror}. Unlike optical sensors that capture diffuse reflections, mmWave signals primarily undergo mirror-like (specular) reflections (Fig.\ref{fig:teaser}(a.1)), resulting in only a limited visible region (Fig.\ref{fig:teaser}(a.2)).
This means that when the sensor transmits a signal toward a surface, such as a wall, the signal reflects away at an angle rather than scattering in all directions. If the reflected signals do not return back to the sensor, the surface remains undetected Fig.~\ref{fig:teaser} (a.2), leading to incomplete and ambiguous reconstructions~\cite{ding2023hidden, hao2024bootstrapping, huang2024dart, wu2024sparseradnet, wang2025high, li2024adapkc, haoradar, yue2025roburcdet, kim2024crt}. 

To overcome this challenge, we introduce \name, a novel benchmark and system that leverages human mobility to enable layout reconstruction and object detection using a single static mmWave sensor. When a person moves within the environment, their presence introduces \textit{multipath effects}, where mmWave wireless signals undergo secondary reflections\footnote{These are in addition to the direct reflections described earlier} before reaching the receiver, as in Fig.\ref{fig:teaser} (b.1). To understand this better, let's consider a simple scenario shown in Fig.\ref{fig:teaser2}. In addition to direct reflections from humans, the radar can also capture signals from higher-order reflection paths (e.g. path 1: radar $\rightarrow$ reflector $\rightarrow$ human $\rightarrow$ radar, and path 2: radar $\rightarrow$ human $\rightarrow$ reflector $\rightarrow$ radar.). These signals create multipath \textit{ghost targets} that move in sync with the human (Ghost 1 and Ghost 2 in Fig.\ref{fig:teaser2})~\cite{liu2024data, liang2024direct, hu2024multipath, li2024belief, valecha2024angle, kim2024multipath}. While these multipath ghosts are traditionally treated as a noise source \cite{park2023multipath, koutsoupidou2019study}, we exploit their geometric properties to infer environmental structures. By analyzing the temporal evolution of multipath reflections, \name\ estimates a rich set of reflectors (blue dots in Fig.\ref{fig:teaser}(b.2)) that encode information about the surrounding layout. \name\ then enhances this reconstruction through a hierarchical diffusion-based optimization model~\cite{ho2020denoising, rombach2022high, chen2023diffusiondet, chen2023rf, chi2024rf}, filling in the missing details and generating the final reconstruction output (red line in Fig.\ref{fig:teaser}(b.2)). Contributions:

\begin{itemize}
\item \textbf{A new benchmark for radar-based indoor scene understanding.}
We collect the first labeled dataset and system for indoor layout understanding using only a \textit{single static mmWave radar}. This dataset provides annotations for structural layouts and object detection, enabling systematic study of radar perception for spatial reasoning.
\item \textbf{A Bi-Angular Multipath Enhancement (BAME) module.}  
We introduce a novel signal enhancement technique that exploits the angular diversity of multipath reflections to reveal “ghost” signals corresponding to occluded walls and structures. The BAME module significantly amplifies these weak yet geometrically meaningful signals, offering a strong initial prior for layout reasoning.
\item \textbf{A Simulation-to-Reality Hierarchical Diffusion (SRHD).}  
To achieve complete scene interpretation—including both wall layout reconstruction and object detection—we develop a radar simulation engine and train a diffusion-based generative model upon it. This approach bridges the gap between synthetic multipath radar signals and real-world measurements, substantially improving the completeness and accuracy of indoor layout recovery.
\end{itemize}

To rigorously evaluate our approach, we deploy the proposed system on a TI MMWCAS-RF-EVM Cascade mmWave radar and collect over \textbf{100 real-world trajectories} encompassing approximately \textbf{50,000 frames}.
Quantitatively, \textbf{\name{}} achieves a mean Chamfer distance of \textbf{16 cm} for wall-layout reconstruction and an \textbf{IoU of 58 \%} for furniture detection—compared to the state-of-the-art baseline~\cite{chen2023environment}, which attains only \textbf{40 cm} Chamfer error and lacks furniture detection capability.
These results establish \textbf{\name{}} as the first comprehensive framework for mmWave-based indoor scene reconstruction, marking significant \textit{qualitative and quantitative} advances in radar layout understanding.

\begin{figure*}[t!]
  \centering
  \includegraphics[width=\linewidth, clip]{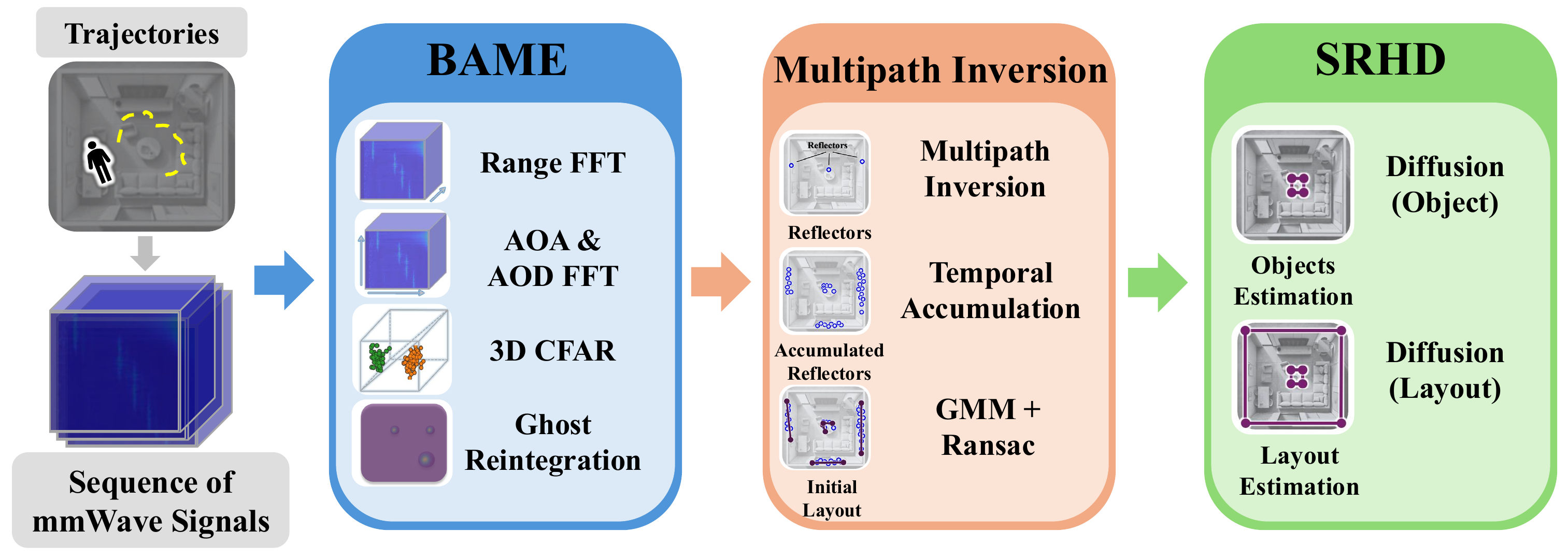}
  \vspace{-0.8cm}
  \caption{\textbf{Pipeline.}  
    Given a sequence of mmWave signals, \name{} first applies the Bi-Angular Multipath Enhancement (BAME) module to each frame to recover previously suppressed ghost paths. These enhanced observations are then processed by our multipath inversion module to estimate the underlying reflector geometry for every frame. By aggregating reflector estimates across the entire trajectory, \name{} forms an initial, geometry-aware layout hypothesis. This coarse reconstruction is subsequently refined by our Sim2Real Hierarchical Diffusion (SRHD) model, which transforms fragmented radar observations into complete wall layouts and object masks.}
  \label{fig:pipeline}
\vspace{-0.40cm}
\end{figure*}

\section{Related Work}

\subsection{Optical-Based Scene Understanding}
Optical methods, including RGB cameras, LiDAR, and their fusion, have been extensively used for scene understanding. 
Recent RGB-based approaches enable accurate object recognition~\cite{girshick2015fast, redmon2016you,liu2016ssd, lin2017focal, carion2020end} and semantic segmentation~\cite{kirillov2023segment,zhao2017pyramid, long2015fully, ronneberger2015u} with the help of deep learning, while LiDAR-based methods provide precise depth information for 3D understanding \cite{xu2019depth, cheng20113d,qi2017pointnet, qi2017pointnet++, wu2018squeezeseg,zhou2026page,zhou2025manydepth2,zhou2023dynpoint}. 
Combining RGB and LiDAR enhances spatial and semantic understanding, benefiting applications such as robotics and augmented reality~\cite{furst2021hperl,asvadi2018multimodal,zia2017rgb,xu2018pointfusion, liu2023bevfusion}. 
Other optical-based methods rely on ambient lighting~\cite{velten2013femto} or laser~\cite{bouman2017turning} to reconstruct the scene around the corner.
However, all optical methods share limitations, including privacy concerns and occlusion challenges \cite{zhang2020object, schneider2010fusing, lin2016iot}.

\subsection{Wireless Signal-Based Scene Understanding}
Past research utilizing wireless signals for scene reconstruction can be broadly categorized into two approaches. The first approach leverages Wi-Fi signals, where the system reconstructs the environment by analyzing reflections captured from multiple vantage points \cite{zhang2021building, yu2024blr}. 
However, these approaches either suffer from low resolution or have deployment overhead.
The second category relies on mmWave signals, where a mmWave radar is mounted on a mobile agent (i.e. robot) to actively scan the environment and perform reconstruction~\cite{yataka2024retr, lu2020see,guidi2015personal,narayanan2020lumos5g,yassin2018mosaic,zeng2024dense,qian20203d,rahman2024mmvr}.  
While this approach has proven effective in providing accurate scene reconstruction, the requirement for having a mobile radar restricts its applicability. 
Recent work has demonstrated the possibility of using a static mmWave radar and multipath for layout reconstruction~\cite{chen2023environment}. However, this method has limitations in fully reconstructing the scene due to blind spots in its field of view and inconsistencies in ghost visibility. Moreover, it is primarily effective for reconstructing large surfaces, such as walls, while lacking the ability to identify or differentiate smaller objects like furniture.
In contrast, \name\ addresses these challenges by leveraging a Bi-Angular Multipath Enhancement module and a Sim2Real Hierarchical Diffusion framework, effectively mitigating blind spot and ghost visibility issues. Additionally, \name\ extends reconstruction capabilities beyond walls, enabling the detection of smaller objects within the scene.

\section{Benchmark}
\label{sec:benchmark}

\paragraph{Dataset Collection.}
To enable standardized evaluation and comparison, we establish the \textbf{\name{}-Indoor Benchmark}, the first dataset dedicated to \textbf{indoor scene understanding using a single static mmWave radar}.
All data was collected using a TI MMWCAS-RF-EVM Cascade mmWave sensor deployed at a fixed height of 1.2\,m, synchronized with an Intel~Realsense depth camera for ground-truth capture.
The benchmark contains over \textbf{100 human motion trajectories}, each approximately 30\,s long, resulting in \textbf{50,000 radar frames} at 20\,Hz.
Data were gathered across \textbf{11 diverse indoor environments}---including offices, corridors, laboratories, and furnished living rooms---under varying furniture arrangements and wall geometries.
To ensure coverage diversity, five volunteers walked along random trajectories, producing rich multipath reflections from multiple incidence and reflection angles.

\noindent \textbf{Annotation Protocol.}
Each radar frame is temporally aligned with its corresponding depth frame via extrinsic calibration and manual verification.
We provide two levels of annotation:
\textbf{(i) Structural Layout Annotations.} 
Wall boundaries and room contours are manually labeled on top-down projections of the depth data and cross-checked with physical floor-plan measurements.
Each frame therefore includes a ground-truth \textit{layout polygon} representing the wall geometry.
\textbf{(ii) Object Annotations.}
Items such as tables, cabinets, and sofas are annotated as axis-aligned bounding boxes in the same 2D coordinate system.
These annotations serve as reference for the object detection.

\noindent \textbf{Evaluation Metrics.}
The benchmark defines two primary tasks: \textit{layout reconstruction} and \textit{object detection}.
\noindent\textit{Layout Reconstruction.}
We report:
Chamfer Distance between predicted and ground-truth wall points, reflecting geometric accuracy.
F1-score at a 15\,cm tolerance threshold to measure boundary completeness.
\noindent\textit{Object Detection.}
We use:
Intersection-over-Union(IoU) between predicted and ground-truth bounding boxes.
Dice Coefficient to quantify segmentation consistency.

\section{Methodology}
In this paper, we introduce \name{}, a novel system that leverages mmWave multipath reflections for accurate indoor layout reconstruction and furniture detection.
Section~\ref{sec:preliminary} introduces the preliminary concepts underlying our approach.
Section~\ref{sec:multipath} provides a detailed geometric analysis for multipath-based layout understanding.
Section~\ref{sec:bmgd} investigates the visibility inconsistency of multipath ghosts and proposes the \textbf{Bi-Angular Multipath Enhancement (BAME)} module to enhance ghost signal observability.
Finally, Section~\ref{sec:sim_diffusion} presents a \textbf{Sim2Real Hierarchical Diffusion (SRHD)}, which generates synthetic training data and employs a hierarchical diffusion model to reconstruct accurate indoor layouts and detect furniture from radar observations. The whole pipeline is shown in Fig.~\ref{fig:pipeline}.

\subsection{Preliminaries}\label{sec:preliminary}

\textbf{RF Signal.}  
An RF signal's phase depends on travel distance; sampling captures its amplitude and phase~\cite{adib2015capturing}. The sampled signal is
\begin{equation}
s = A e^{-j 2\pi r / \lambda}
\end{equation}
where \( r \) is the traveled distance, \( \lambda \) is the wavelength, and \( A \) is the amplitude.

\textbf{Angle Estimation.} 
For an \( N \)-element antenna array, the received signal power from direction \( \theta \) can be estimated as:
\begin{equation}
P(\theta) = \sum_{n=1}^{N} s_n e^{j 2\pi (n d \cos \theta) / \lambda}
\label{equa:angle}
\end{equation}
where \( s_n \) is the received signal at the \( n \)-th antenna, \( d \) is the spacing between antennas, and \( \lambda \) is the wavelength~\cite{adib2015capturing}.  
In practice we use the standard sum-of-phasors/beamforming formulation (Eq.~\ref{equa:angle}) to build range--angle heatmaps which serve as the primary observation for ghost identification and later geometric inference.

\begin{figure}[t!]
    \centering
    \includegraphics[width=\linewidth, clip]{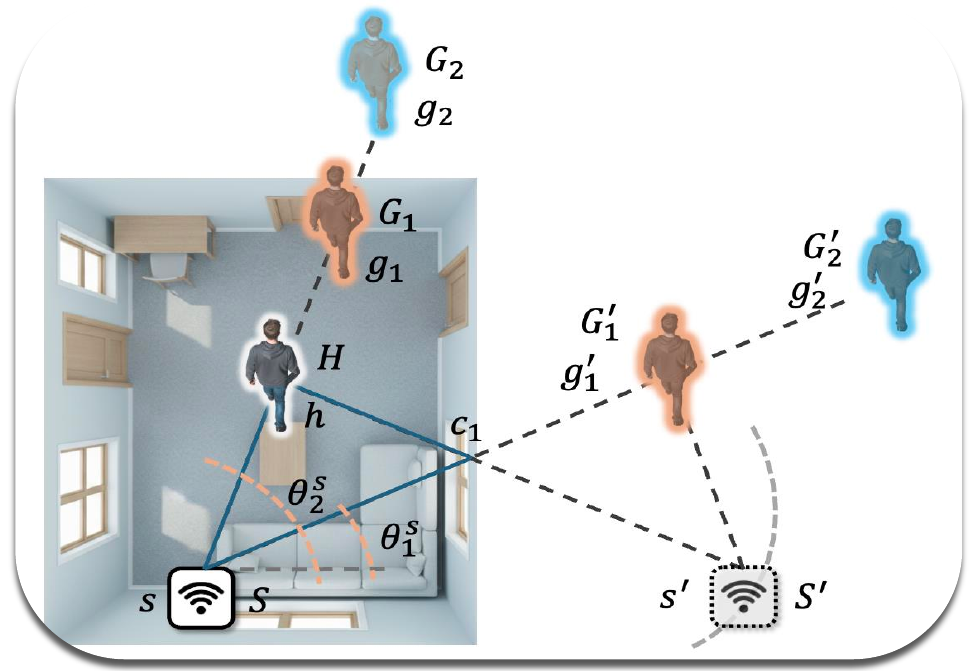}   
    \vspace{-0.3cm}
    \caption{\textbf{Multipth Inversion.}  
    Geometric relationships between ghost targets and reflectors. }
\label{fig:posterior}
\vspace{-0.40cm}
\end{figure}
\vspace{-0.20cm}

\subsection{Multipath Inversion}\label{sec:multipath}

\textbf{Ghost Target Formation and Identification.}  
As shown in Fig.~\ref{fig:posterior}, when a static radar sensor \(S\) observes a moving human target \(H\), additional reflections—\textit{ghost targets}—appear due to multipath propagation.  
We focus on the first- and second-order ghosts, denoted as \(G_1 (\text{following}\ s\rightarrow c_1 \rightarrow h \rightarrow s), G'_1 (\text{following}\ s\rightarrow h \rightarrow c_1 \rightarrow s), G_2(\text{following}\ s\rightarrow h \rightarrow c_1 \rightarrow h \rightarrow s)\) and \(G'_2(\text{following}\ s\rightarrow c_1 \rightarrow h \rightarrow c_1 \rightarrow s)\), corresponding to two- and three-bounce reflections~\cite{wu2023nlos}.  
Given the rapid attenuation of higher-order paths, only these four ghosts are considered. 
Their visibility in the range--angle map depends on scene geometry and signal processing.

We begin by using the Constant False Alarm Rate (CFAR) detection method, followed by RANSAC clustering to identify several high power regions ("clusters") within the mmWave image. Then, we can identify which of these clusters correspond to different ghost targets in four steps:
\(H\): select clusters with \(m_i > \tau\max(m_i)\) where $m_i$ is the received magnitude of the i\textsuperscript{th} cluster and choose the one with the smallest range.  
\textbf{\(G_1\):} same direction as \(H\) but slightly larger range.  
\textbf{\(G'_1\):} same range as \(G_1\) but different direction.  
\textbf{\(G'_2\):} larger range than \(G'_1\) and aligned with \(\vec{sg'_1}\).
Two points are considered identical if their range difference is below 15\,cm or angular difference below 15°, consistent with real-world spatial resolution.  
Detailed thresholds and pseudocode are provided in the Appendix.


\textbf{Reflector Point Estimation.}  
The reflector point corresponding to \(G'_1\) can be estimated as:
\begin{equation}
    sc_1 = \frac{2|\vec{sg'_1}|^2 - 2|\vec{sg'_1}||\vec{sh}|}{2|\vec{sg'_1}| - |\vec{sh}| \cos (\theta^s_2 - \theta^s_1) - |\vec{sh}|}
\label{Equ:ghost1}
\end{equation}
The proof can be found in the appendix. 
We first apply a Gaussian Mixture Model (GMM) to group the estimated reflector points into spatially coherent clusters, and then use RANSAC line fitting to infer the dominant line structure.

\subsection{Bi-Angular Multipath Enhancement (BAME)}
\label{sec:bmgd}

\begin{figure}[t!]
  \centering
  \includegraphics[width=\linewidth, clip]{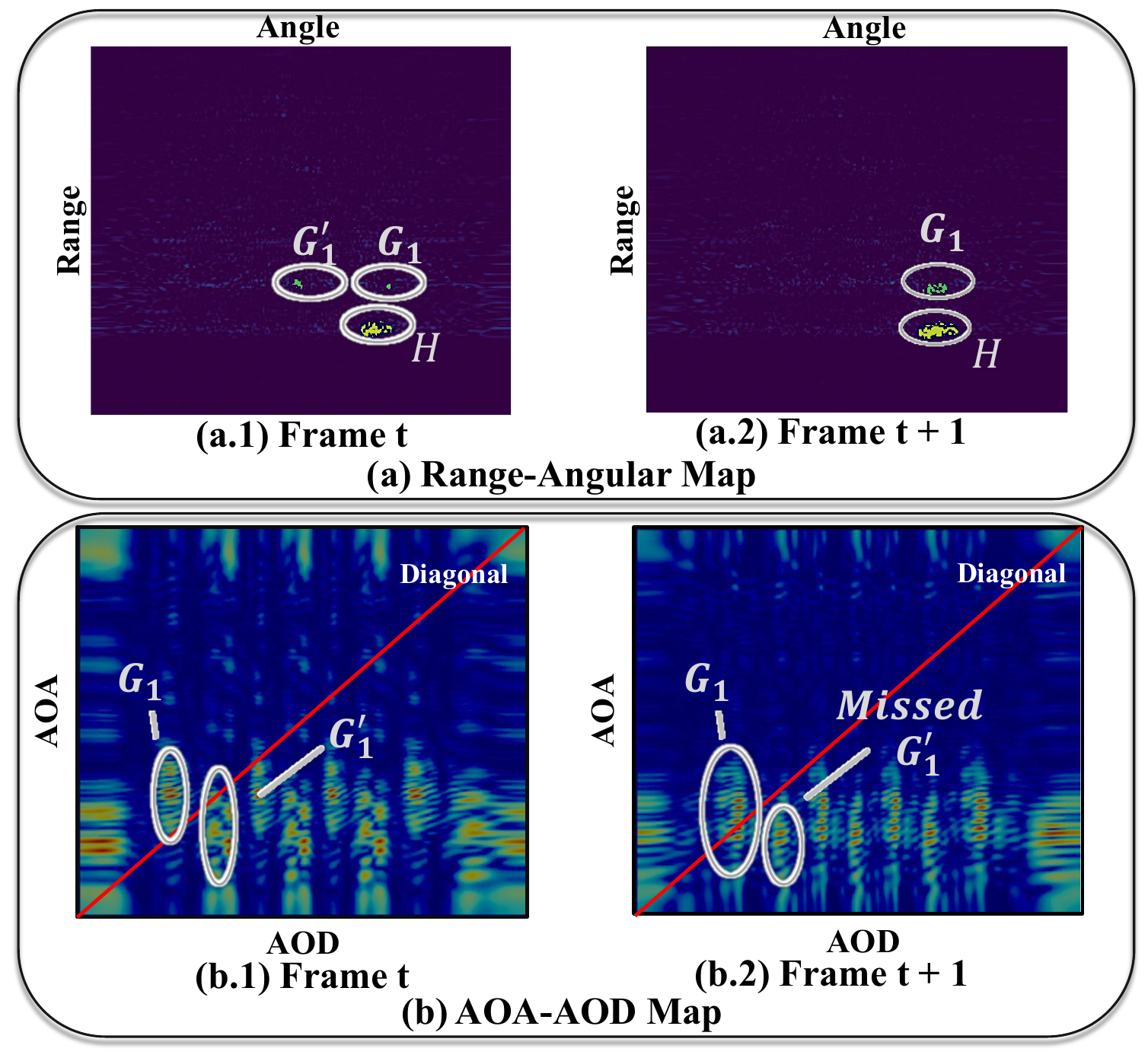}
  \vspace{-0.8cm}
  \caption{\textbf{Bi-Angular Multipath Enhancement (BAME).}  
    Figure~(a.1) shows a case where the ghost targets can be recovered using conventional beamforming, and Fig. (a.2) shows a case where the ghost target $G'_1$ cannot be recovered using conventional beamforming.
    Figure~(b) illustrates the corresponding AOA--AOD response at the range of the ghost targets $G'_1$ and $G_1$.  
    Notably, $G'_1$ disappears in the standard range--angle map (Fig.~(a.2)) because its AOA and AOD are not aligned with the diagonal, causing it to be suppressed by the $\text{AOA}=\text{AOD}$ assumption.}
  \label{fig:enhance3d}
\vspace{-0.40cm}
\end{figure}

Accurately identifying ghost targets is essential for reconstructing reliable reflector geometry.  
However, during experiments, we found that ghost visibility changes drastically across frames, as shown in Fig.~\ref{fig:enhance3d}(a). To increase the ghost detection capability, we propose algorithm named Bi-Angular Multipath Enhancement (BAME).

\textbf{Motivation.}
The main reason lies in the difference between the \textbf{Angle of Arrival (AOA)} and \textbf{Angle of Departure (AOD)} of the reflected paths.  
For example, as shown in Fig.~\ref{fig:posterior}, ghost $G'_1$ has AOA $= \theta^s_1$ and AOD $= \theta^s_2$.  
This asymmetry is crucial because most conventional radar beamforming formulations (Eq.~\ref{equa:angle}) implicitly assume $\text{AOA} = \text{AOD}$, causing reflections with mismatched AOA and AOD---such as $G'_1$---to be suppressed or disappear.

\textbf{Mathematical insight.}
If we separate the processing of AOA and AOD information, we obtain a more general formulation to find the received power at a given AOA, $\theta_{\text{AOA}}$, AOD, $\theta_{\text{AOD}}$, and range, $r$:
\begin{equation}
\label{eq:biangular}
\begin{aligned}
S&(\theta_{\text{AOA}}, \theta_{\text{AOD}},r) 
= \\
&\sum_{i=1}^{N_r}
\left(
    \sum_{k=1}^{N_t}
        s_{ikr}\, e^{j2\pi (k d_t \cos\theta_{\text{AOD}})/\lambda}
\right)  
e^{j2\pi (i d_r \cos\theta_{\text{AOA}})/\lambda}.
\end{aligned}
\end{equation}

\noindent where $N_r$ and $N_t$ are the number of receivers and transmitters, and $d_r$ and $d_t$ are the spacing between successive receivers and transmitters, respectively. $s_{ikr}$ is the signal from the k\textsuperscript{th} transmitter to the i\textsuperscript{th} receiver, at a fixed range $r$ (after an FFT along the range dimension). Here, the inner summation encodes the AOD contribution and the outer summation represents the AOA component.  
The corresponding 2D response map (at a fixed range $r$) is illustrated in Fig.~\ref{fig:enhance3d}(b).
However, in standard MIMO radar processing, the receiver and transmitter indices 
$\{r_i\}_{i=1}^{N_r}$ and $\{t_j\}_{j=1}^{N_t}$ are combined into a single 
``virtual array'' index, such that Eq.~\ref{equa:angle} becomes:
\begin{equation}
S(\theta, r) =
\sum_{i=1}^{N_r} \sum_{k=1}^{N_t}
s_{ikr}\, e^{j2\pi (d_r i + d_t k)\cos\theta / \lambda}.
\end{equation}
This formulation implicitly merges AOA and AOD into a single equivalent angular dimension, 
effectively assuming $\text{AOA} \approx \text{AOD}$.  
While this simplification is adequate for direct reflections, it suppresses multipath components 
where the two angles differ.  
Thus, first-order ghosts such as $G'_1$---which possess distinct AOA and AOD values---are 
partially or completely invisible in the resulting range--angle map.

\textbf{Empirical observation.}
As illustrated in Fig.~\ref{fig:enhance3d}(a), ghost $G'_1$ is visible in frame $t$ but disappears in frame $t{+}1$.  
In frame $t$, both $G_1$ and $G'_1$ lie along the diagonal of the Range--AOA--AOD map (left of Fig.~\ref{fig:enhance3d}(b)), making them detectable.  
However, in frame $t{+}1$, only $G_1$ remains on the diagonal, while $G'_1$ shifts off-diagonal and consequently vanishes from the range--angle response (right of Fig.~\ref{fig:enhance3d}(b)).  
This temporal inconsistency demonstrates that conventional diagonal-only beamforming fails to capture multipath components with distinct AOA--AOD pairs.

\textbf{Proposed Method.}
To overcome the above limitation, we introduce \textbf{Bi-Angular Multipath Enhancement (BAME)}, 
a two-angle radar processing pipeline that models both AOA and AOD dimensions.  
BAME reconstructs the full Range--AOA--AOD cube through the follows:

\begin{enumerate}
    \item \textbf{Range FFT:} Perform a fast Fourier transform along the range dimension to obtain the distance spectrum.  
    \item \textbf{Bi-Angular Beamforming:} Apply Eq.~\ref{eq:biangular} separately to receiver and transmitter dimensions to compute independent AOA and AOD responses.  
    \item \textbf{3D CFAR Detection:} Detect high-energy clusters within the Range--AOA--AOD cube using a three-dimensional CFAR filter.  
    \item \textbf{Ghost Reintegration:} For clusters exhibiting distinct AOA and AOD values, reinsert them into the range--angle map as valid ghost reflections.  
\end{enumerate}

\textbf{Effect.}
This bi-angular formulation recovers the missing off-diagonal reflections that conventional processing suppresses, 
enabling consistent detection of both $G_1$ and $G'_1$.  
Consequently, the reconstructed range--angle map becomes more complete and geometrically faithful, 
forming a robust foundation for subsequent layout reconstruction.
\subsection{Sim2Real Hierarchical Diffusion (SRHD)}
\label{sec:sim_diffusion}

Using the outputs of the Bi-Angular Multipath Enhancement (BAME) and Multipath Inversion modules, we obtain an initial reconstruction of the indoor layout. 
While this estimate captures the coarse spatial structure, it remains fragmented—comprising disjoint line segments that fail to form continuous walls or provide reliable object detection.
Inspired by the strong reasoning capability of modern diffusion-based generative models across vision and 3D perception, \name{} introduces a \textbf{geometry-aware diffusion framework} that bridges the gap between fragmented radar observations and complete indoor layout reconstruction. 
However, training such a model requires a large corpus of radar-to-layout correspondences, which is unavailable due to the novelty of this sensing modality. 
To overcome this data scarcity, we design a high-fidelity \textbf{layout simulation engine} and a \textbf{hierarchical diffusion architecture} that jointly enable robust, geometry-consistent reconstruction from partial and noisy radar observations.

\begin{figure}[t!]
  \centering
  \includegraphics[width=\linewidth, clip]{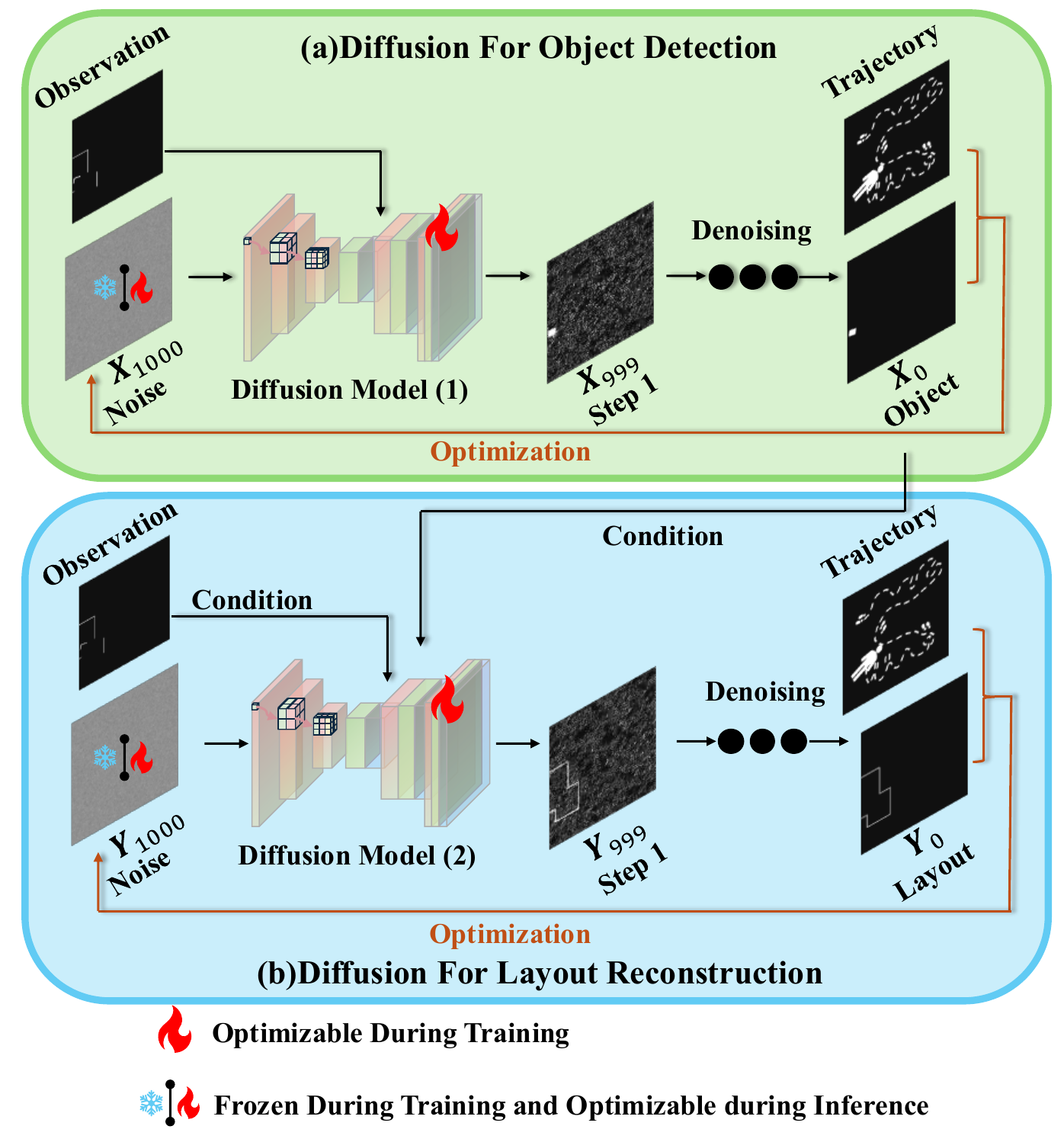}
  \vspace{-0.8cm}
  \caption{\textbf{Sim2Real Hierarchical Diffusion (SRHD).}  In the first stage, \name{} predicts the furniture-based segmentation map. In the second stage, \name{} estimates the layout-based segmentation map.}
  \label{fig:generative}
\vspace{-0.40cm}
\end{figure}

\textbf{Layout Simulation Engine.}
We construct a simulator based on a corpus of 35,000 real-world indoor floor plans. 
Each layout is first converted into a 2D structural skeleton, and a virtual radar sensor is randomly placed at arbitrary positions and orientations. 
Multiple rays are then emitted from the radar, and only the first intersection point per ray is retained, simulating the physical propagation property of mmWave signals—where only the nearest surface contributes to the reflection. 
To emulate real-world multipath interactions, we insert $N$ randomly positioned bounding boxes as objects and select one reflective line segment per box to model stochastic material-dependent scattering.

\begin{figure}[t!]
    \centering
    \includegraphics[width=\linewidth, clip]{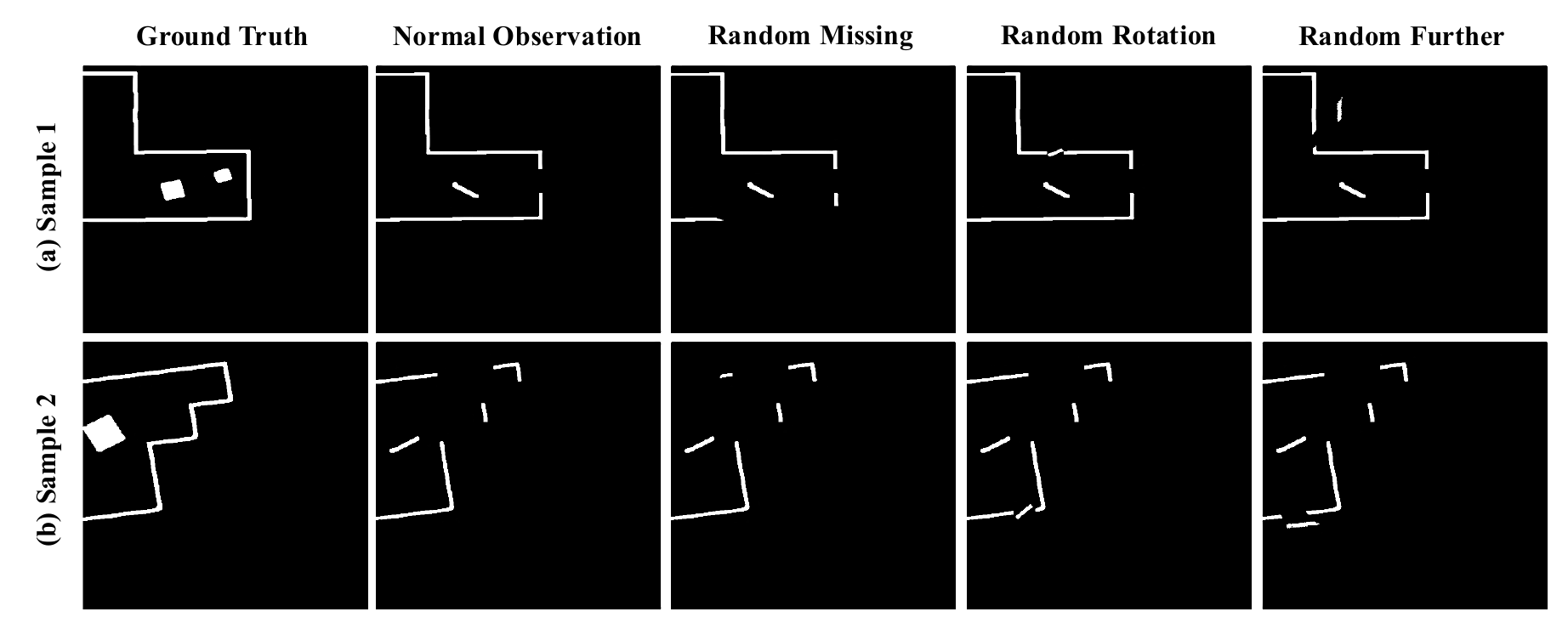}   
    \vspace{-0.7cm}
    \caption{\textbf{Illustration of Simulator and Data Augmentation.} The first column represents the ground truth in the simulator, while the second column shows partial observations resulting from occlusion. The third column depicts randomly missing regions, the fourth column illustrates random rotations, and the fifth column demonstrates random scaling.  }
\label{fig:simulator}
\vspace{-0.5cm}
\end{figure}

\textbf{Data Augmentation.}
To enhance generalization from simulation to reality, we apply three augmentation strategies (Fig.~\ref{fig:simulator} Appendix.) that inject structural noise:
\begin{itemize}
    \item \textbf{Random Missing.}
    Randomly remove line segments within selected angular intervals 
    $\Theta_{\text{del}}$
    to simulate occluded or incomplete radar coverage.
    \item \textbf{Random Rotation.}
    Rotate all foreground pixels $(x_p,y_p)$ within a chosen angular range $\Theta_{\text{rot}}$ by angle $\alpha_i$ around the radar’s center $(x_c, y_c)$:
    \begin{equation}
        \begin{bmatrix} x_p' \\ y_p' \end{bmatrix} = 
        \begin{bmatrix} \cos\alpha_i & -\sin\alpha_i \\ \sin\alpha_i & \cos\alpha_i \end{bmatrix} 
        \begin{bmatrix} x_p - x_c \\ y_p - y_c \end{bmatrix} + 
        \begin{bmatrix} x_c \\ y_c \end{bmatrix}.
    \end{equation}
    This augmentation mimics sensor pose errors and small orientation drifts.
    \item \textbf{Random Scaling.}
    Scale each foreground pixel $(x_j, y_j)$ radially relative to the radar position $(x_r, y_r)$ by a random factor $a_i$:
    \begin{equation}
        \begin{bmatrix} x_p' \\ y_p' \end{bmatrix} = 
        \begin{bmatrix} x_r \\ y_r \end{bmatrix} 
        + a_i 
        \left( 
        \begin{bmatrix} x_p \\ y_p \end{bmatrix} - 
        \begin{bmatrix} x_r \\ y_r \end{bmatrix} 
        \right),
    \end{equation}
    which simulates depth uncertainty and propagation delay variations.
\end{itemize}
These augmentations improve robustness by exposing the model to spatial noise and non-ideal radar conditions.

\textbf{Hierarchical Diffusion Model.}
With the augmented dataset, we train a two-stage diffusion model (Fig.~\ref{fig:generative}) that reconstructs both semantic and geometric completeness:
\begin{itemize}
    \item \textbf{Stage 1: Object Detection.}
    Given a partial observation $\mathbf{O}$, the first diffusion model $f_1$ predicts a binary object detection map 
    $\mathbf{X}_0 \in \{0,1\}^{H \times W}$:
    \begin{equation}
        \mathbf{X}_0 = f_1(\mathbf{O}, \mathbf{X}_{1000}; \theta_1),
    \end{equation}
    where $\mathbf{X}_{1000}$ denotes Gaussian noise.
    This step provides semantic priors that guide reasoning in the next stage.
    \item \textbf{Stage 2: Wall Structure Prediction.}
    The second model $f_2$ takes both $\mathbf{O}$ and $\mathbf{X}_{1000}$ as inputs to reconstruct the full wall configuration $\mathbf{Y}_{1000}$:
    \begin{equation}
        \mathbf{Y}_{0} = f_2(\mathbf{O}, \mathbf{Y}_{1000}, \mathbf{X}_0; \theta_2).
    \end{equation}
    A channel-attention fusion module integrates geometric and semantic cues, enforcing spatial coherence.
\end{itemize}

\textbf{Spatial Consistency Optimization.}
During inference, we further refine the latent noise variables $\mathbf{X}_{1000}$ and $\mathbf{Y}_{1000}$ using an overlap-based optimization that enforces plausibility between human motion and generated structures:
\begin{equation}
\begin{aligned}
\mathcal{L}_{\text{overlap}} 
= \sum_{(x_t, y_t) \in \mathcal{T}}
\Big[
    \mathds{1}[(x_t, y_t) \in \mathcal{W}] 
 +\,
    \mathds{1}[(x_t, y_t) \in \mathcal{B}]
\Big],
\end{aligned}
\end{equation}
where $\mathcal{T}$ denotes the trajectory of human motion, and $\mathcal{W}$ and $\mathcal{B}$ represent wall and object masks, respectively. 
\begin{table*}[h]
    \centering
    \resizebox{\linewidth}{!}{
    \begin{tabular}{llccccccccccccc}
        \hline
        method & Metric & Type & S1 & S2 & S3 & S4 & S5 & S6 & S7 & S8 & S9 & S10 & S11 & AVG\\
        \hline
        \multirow{2}{*}{BRL~\cite{hao2024bootstrapping}} 
         & \cellcolor{col2} IoU (\%) $\uparrow$ & S & 00.00 & 00.00 & 00.00 & 00.00 & 00.00 & 12.93 & 00.00 & 00.00 & 00.00 & 00.00 & 00.00 & 01.17 \\
         & \cellcolor{col2} Dice (\%) $\uparrow$& S & 00.00 & 00.00 & 00.00 & 00.00 & 00.00 & 20.03 & 00.00 & 00.00 & 00.00 & 00.00 & 00.00 & 01.82 \\
        \multirow{2}{*}{EMT~\cite{chen2023environment}+SRHD} 
         & \cellcolor{col2} IoU (\%) $\uparrow$ & M & 00.00 & 00.00 & 15.61 & 00.00 & 00.00 & 25.77 & 23.44 & 19.93 & 00.00 & 00.00 & 00.00 & 07.70 \\
         & \cellcolor{col2} Dice (\%) $\uparrow$& M& 00.00 & 00.00 & 26.13 & 00.00 & 00.00 & 43.98 & 38.65 & 30.54 & 00.00 & 00.00 & 00.00 & 12.66 \\
        \multirow{2}{*}{\name{}(Ours)} 
         & \cellcolor{col2} IoU (\%) $\uparrow$ & M & 58.70 & 96.02 & 29.48 & 47.90 & 55.24 & 47.12 & 64.70 & 37.91 & 64.43 & 74.19 & 59.87 & 57.78 \\
         & \cellcolor{col2} Dice (\%) $\uparrow$& M & 73.93 & 96.68 & 43.64 & 64.26 & 70.24 & 61.55 & 71.20 & 48.46 & 77.41 & 84.86 & 70.52 & 69.34 \\
        \hline
    \end{tabular}}
    \vspace{-0.25cm}
    \caption{\textbf{Object Detection Results:} This table presents the object detection results of our method and baselines, evaluated using Intersection over Union (IoU) and Dice coefficient metrics. "S" denotes single-frame input, and "M" denotes multi-frame input.}
    \label{tab:iou_dice}
    \vspace{-0.5cm}
\end{table*}
Here, $\mathds{1}[\cdot]$ is the indicator function that equals $1$ if a trajectory point collides with an obstacle and $0$ otherwise. 
Minimizing $\mathcal{L}_{\text{overlap}}$ iteratively adjusts the latent noise inputs, eliminating wall–trajectory and object–trajectory intersections. 
This reverse optimization ensures that the final reconstruction is physically valid—walls remain static and non-penetrable, while human trajectories traverse only feasible free-space regions. A discussion is presented in our Appendix.

\section{Experimental} 

\subsection{Experimental Setting}
\textbf{Tasks.} 
We evaluate \name{} on two core tasks: \emph{layout reconstruction} and \emph{object detection}. 
For layout reconstruction, we report Chamfer Distance to quantify geometric accuracy and F1-score (15\,cm tolerance) to measure boundary completeness. 
For object detection, we adopt standard 2D metrics: Intersection-over-Union (IoU) and the Dice coefficient.
\textbf{Baselines.}
Since \name{} is the first system to perform \emph{both} layout reconstruction and object detection using a \emph{single static mmWave radar}, no prior method directly matches our setting. 
For object detection, we compare against:  
(1) the state-of-the-art learning-based single-radar detector BRL~\cite{hao2024bootstrapping};  
(2) the multi-frame radar layout reconstruction method EMT~\cite{chen2023environment}.  
Because EMT does not support object detection, we augment its outputs using the same post-processing used in \name{} to provide a fair comparison. We rename it as EMT~\cite{chen2023environment} + SRHD.
For layout reconstruction, we directly compare against EMT~\cite{chen2023environment}, a strong existing multi-frame radar layout reconstruction baseline.

\subsection{Object Detection}  
Tab.~\ref{tab:iou_dice} and Fig.~\ref{fig:demo} present the object detection results. 
As shown in Fig.~\ref{fig:demo}, the baseline method (EMT~\cite{chen2023environment}) fails to recover reflectors from small objects, making it unable to produce meaningful object detections.
To enable a fair comparison, we report two additional baselines in Tab.~\ref{tab:iou_dice}: 
(1) \textbf{EMT~\cite{chen2023environment} + SRHD}, where we apply our Sim2Real Hierarchical Diffusion module on top of the EMT output, and 
(2) \textbf{BRL}~\cite{hao2024bootstrapping}, a deep-learning–based single-radar object detector.
Across all settings, \name{} delivers substantially stronger performance. 
By combining Ghost Signal Enhancement with Sim2Real Hierarchical Diffusion, \name{} reliably localizes objects and generalizes to complex indoor scenes. 
\name{} achieves an \textbf{IoU of 57.78} and a \textbf{Dice score of 69.34}, far surpassing the best baseline (IoU = 7.70, Dice = 12.66).

\subsection{Layout Reconstruction}  
\begin{figure}[t!]
    \centering
    \includegraphics[width=\linewidth, clip]{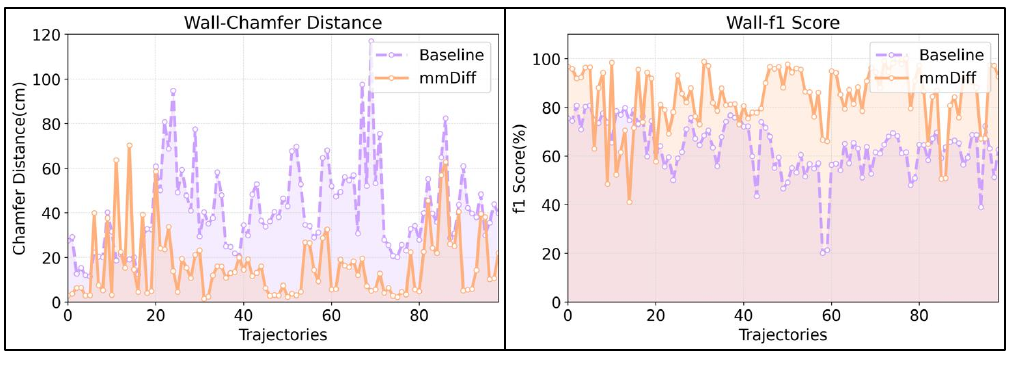}   
    \vspace{-0.7cm}
    \caption{\textbf{Wall Reconstruction Across 100 Trajectories.} Comparison of our method and the baseline (\textbf{EMT}~\cite{chen2023environment}) over 100 real-world trajectories, evaluated using Chamfer Distance and F1-score for wall layout reconstruction.}
    \label{fig:chamfer_scene} \vspace{-0.5cm}
\end{figure}

\begin{figure*}[t!]
  \centering
  \includegraphics[width=\linewidth, clip]{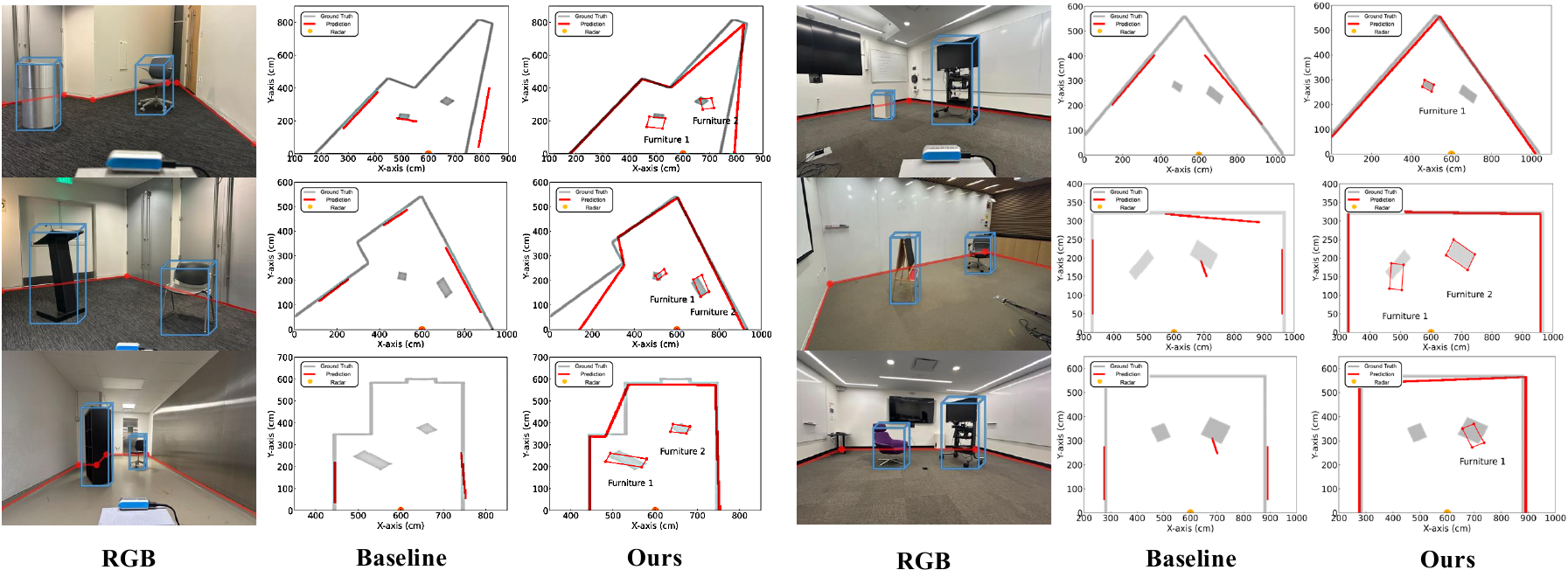}
  \vspace{-0.3cm}
  \caption{\textbf{Comparison Between Our Method and the Baseline.}
The first column shows the RGB reference images of the corresponding scenes.
The second column presents the reconstruction results produced by EMT~\cite{chen2023environment}.
The third column shows the results generated by our method, including both layout reconstruction and object detection.}
  \label{fig:demo}
\vspace{-0.3cm}
\end{figure*}

Fig.~\ref{fig:demo} also presents qualitative comparisons for layout reconstruction between our method and the baseline \textbf{EMT}~\cite{chen2023environment}. 
Since EMT~\cite{chen2023environment} is inherently capable of wall-layout reconstruction, we evaluate it in its original form without combining it with SRHD.
We further summarize quantitative results across 100 trajectories in Fig.~\ref{fig:chamfer_scene}. 
Compared with EMT~\cite{chen2023environment}, \name{} produces substantially more complete and structurally coherent layouts, particularly in regions where multipath coverage is sparse.
Quantitatively, \name{} achieves a markedly lower average Chamfer distance of \textbf{16.03 cm} (vs.\ \textbf{39.06 cm} for EMT) and a significantly higher F1-score of \textbf{83.63} (vs.\ \textbf{63.43}). 
These results demonstrate the effectiveness of our Bi-Angular Multipath Ghost Enhancement and Sim2Real Hierarchical Diffusion framework in recovering both global room geometry and fine-grained structural details.

\subsection{Effect of Trajectory Length}  
\begin{figure}[t!]
    \centering
    \includegraphics[width=\linewidth, clip]{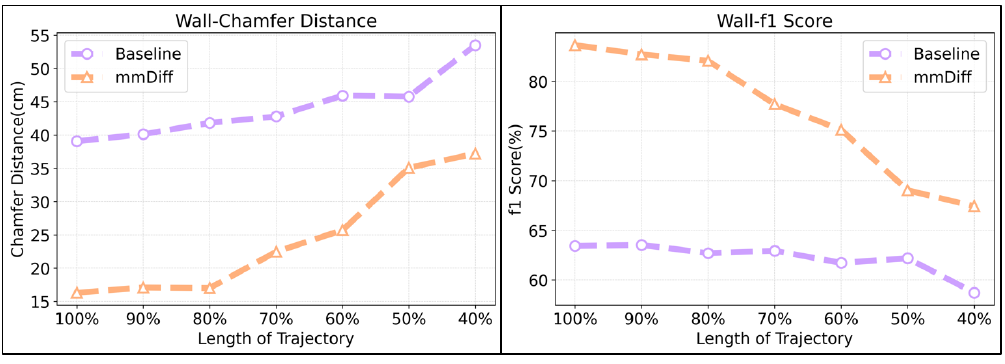}   
    \vspace{-0.3cm}
    \caption{\textbf{Results Across Varying Trajectory Lengths.} 
    Comparison between our method and the baseline \textbf{EMT}~\cite{chen2023environment}. 
    The figure reports layout reconstruction performance—Chamfer distance and F1-score—under  shorter human trajectories, illustrating the robustness of \name{} when input coverage becomes limited.}
    \label{fig:chamfer_traj}
    \vspace{-0.3cm}
\end{figure}

Our method employs 30-second trajectories for layout reconstruction. To assess the impact of trajectory length on performance, we analyze the results as the trajectory length decreases shown in Fig.~\ref{fig:chamfer_traj}. Both our method and the baselines (\textbf{EMT}~\cite{chen2023environment}) exhibit performance degradation as the trajectory length shortens. However, the decline in F1-score is more pronounced in our method. This is attributed to the nature of the F1-score computation, which takes into account predictions within a specific error threshold. As the trajectory length decreases, the diffusion model generates more diverse outputs due to the reduced input information, which consequently has a greater impact on the F1-score. Notably, even with 40\% of the full-length trajectories, \name{} still achieves a lower Chamfer distance compared to the baseline performance using the full-length trajectories.

\begin{table}[h]
    \centering
    \vspace{-0.1cm}
    \resizebox{0.88\linewidth}{!}{
    \begin{tabular}{lcc}
        \hline
        Method/Metric & \cellcolor{col2} F1-score $\uparrow$ & \cellcolor{col1} Chamfer $\downarrow$ \\
        \hline
        Baseline & 63.43 \% & 39.06 cm \\
        \name{} w/ G & 73.37 \% & 32.31 cm \\
        \name{} w/ G, D & 78.84 \% & 19.82 cm \\
        \name{} w/ G, D, R & 83.63 \% & 16.32 cm \\
        \hline
    \end{tabular}}
    \vspace{-0.25cm}
    \caption{Ablation study results showing the impact of different components on F1-score and Chamfer Distance.}
    \label{tab:ablation_study}
    \vspace{-0.3cm}
\end{table}

\subsection{Ablation Study}  
To assess the contribution of key components in our framework, we conduct an ablation study, the results of which are summarized in Tab.~\ref{tab:ablation_study}. We evaluate the impact of three critical components: Ghost Signal Enhancement, the diffusion model, and reverse optimization. The configurations tested include: \textbf{Model w/ G}, which includes only Ghost Signal Enhancement; \textbf{Model w/ G, D}, which includes both Ghost Signal Enhancement and the diffusion model; and \textbf{Model w/ G, D, R}, the full model incorporating Ghost Signal Enhancement, the diffusion model, and reverse optimization.

Our findings reveal a progressive improvement in Chamfer distance, which underscores the contribution of each module in enhancing reconstruction accuracy: 1) The inclusion of Ghost Signal Enhancement reduces the Chamfer distance from \textbf{39.06 cm} to \textbf{32.31 cm}, highlighting its ability to improve initial detection by enhancing reflector visibility. 2) The addition of the diffusion model further refines the reconstruction, reducing the Chamfer distance to \textbf{19.82 cm}, demonstrating its ability to generate more accurate layout details by leveraging generative modeling. 3) The incorporation of reverse optimization leads to the final refinement, achieving a Chamfer distance of \textbf{16.32 cm}. This stage fine-tunes the results by optimizing the generative outputs, further improving the quality and precision of the reconstructed layout.
The progressive reduction in Chamfer distance across these configurations demonstrates the complementary nature of each component. Ghost Signal Enhancement aids in initial detection, the diffusion model enhances overall reconstruction, and reverse optimization refines the final output, collectively resulting in a substantial improvement in reconstruction accuracy.

\section{Conclusion}
In conclusion, this paper introduces a novel method for indoor layout reconstruction using a single static mmWave sensor. By leveraging human mobility-induced multipath effects and integrating a generative model, our system, \name{}, effectively tackles challenges such as specularity and incomplete layout reconstruction. Experimental results demonstrate that \name{} achieves a Chamfer distance of 16 cm for layout reconstruction and an IoU of 58\% for furniture detection, outperforming conventional approaches. This work highlights the potential of privacy-preserving mmWave sensing for applications in smart homes, security, and virtual/augmented reality, and lays the groundwork for future improvements in more complex environments, such as multi-person scenarios and shorter trajectories.

\newpage

\section{Acknowledgments.} We thank the anonymous reviewers and the Signal Kinetics group for their help and feedback. This research is sponsored by NSF (Award \#2313234), Amazon MIT Science Hub, and the MIT Media Lab. Laura Dodds is supported by the Amazon Robotics Thriving Stars Fellowship.

{
    \small
    \bibliographystyle{ieeenat_fullname}
    \bibliography{main}
}

\newpage

\section{Experiments}

\textbf{Data Process:} The ground truth is generated using a RealSense depth camera. Specifically, the depth camera captures the 3D point cloud of the surrounding environment. This point cloud is then transformed into the radar coordinate system using pre-calibrated intrinsic and extrinsic matrices.
To create ground truth for wall estimation, we project all 3D points onto the radar's horizontal (XZ) plane by collapsed along Z. This projection provides a reliable reference for evaluating our wall detection algorithm.
For object detection, we first use the depth camera to identify the positions of objects in the scene. Then, using geometric cues from the depth data and simple measurement rules, we estimate the actual dimensions of each object. Based on the estimated position and size, we construct 2D bounding boxes in the radar coordinate system, which serve as ground truth for evaluating object detection performance.

\section{Detail for Section - 4.2. Multipath Inversion}
\subsection{Ghost Target Formation and Identification}
To make the ghost and human target identification process transparent, we provide a clear step-by-step pseudocode description in Algorithm 1.

\begin{algorithm}[t]
\caption{Ghost Target Formation and Identification}
\label{alg:ghost_detection}
\KwIn{
Clusters $\mathcal{C}=\{(r_i,\theta_i,m_i)\}$ obtained from CFAR detection, where $r_i$ denotes the range, $\theta_i$ the angle, and $m_i$ the reflection magnitude.
Threshold ratio $\tau=0.4$; 
Range tolerance $\delta_r=0.15$m; 
Angle tolerance $\delta_\theta=15^\circ$.
}
\KwOut{Human $H$, first-order ghosts $G_1, G_1'$, second-order ghosts $G_2, G_2'$.}

\textbf{Step 1: Identify Human $H$}\;
$M_{\max} \leftarrow \max_i m_i$ \\
$\mathcal{C}_{\text{valid}} \leftarrow \{ c \in \mathcal{C} \mid m_c > \tau M_{\max} \}$ \\
$H \leftarrow \arg\min_{c\in \mathcal{C}_{\text{valid}}} r_c$

\BlankLine
\textbf{Step 2: Identify $G_1$ (same direction, slightly larger range)}\;
$\mathcal{G}_1 \leftarrow \{ c \in \mathcal{C} \mid |\theta_c-\theta_H|<\delta_\theta,\; 0<r_c-r_H<\delta_{r} \}$ \\
$G_1 \leftarrow \arg\min_{c\in\mathcal{G}_1} r_c$ \ if $\mathcal{G}_{1}\neq\emptyset$, else $\emptyset$

\BlankLine
\textbf{Step 3: Identify $G_1'$ (same range, different direction)}\;
$\mathcal{G}_{1'} \leftarrow \{ c \in \mathcal{C} \mid |r_c-r_{G_1}|<\delta_r,~ |\theta_c-\theta_{G_1}| \ge \delta_\theta \}$ \\
$G_1' \leftarrow \arg\max_{c\in\mathcal{G}_{1'}} m_c$ if $\mathcal{G}_{1'}\neq\emptyset$, else $\varnothing$

\BlankLine
\textbf{Step 4 (Optional): Identify $G_2$ (same direction as $H$, farther than $G_1$)}\;
$\mathcal{G}_2 \leftarrow \{c\in\mathcal{C} \mid |\theta_c-\theta_H|<\delta_\theta,~ r_c > r_{G_1} \}$ \\
$G_2 \leftarrow \arg\min_{c\in\mathcal{G}_2} r_c$ if $\mathcal{G}_2\neq\emptyset$, else $\varnothing$

\BlankLine
\textbf{Step 5: Identify $G_2'$ (aligned with $\vec{G_1'}$, farther range)}\;
$\mathcal{G}_{2'} \leftarrow \{c\in\mathcal{C} \mid r_c>r_{G_1'},~ |\theta_c-\theta_{G_1'}|<\delta_\theta\}$ \\
$G_2' \leftarrow \arg\min_{c\in\mathcal{G}_{2'}} r_c$ if $\mathcal{G}_{2'}\neq\emptyset$, else $\varnothing$

\Return{$H, G_1, G_1', G_2, G_2'$}
\end{algorithm}

\subsection{Reflector Point Estimation}

\subsubsection{First-Bounce Ghost}
The first-bounce ghost point, denoted as $G_1'$, is geometrically related to the source $S'$ as follows:  

\begin{equation}
    2 |sg'_1| = |sh| + |hc_1| + |sc_1|,
\end{equation}  

\noindent where $|sg'_1|$ represents the distance from the radar to the first-bounce ghost point, $|sh|$ is the distance from the radar to the human being, $|hc_1|$ is the distance from the human being to the reflector point, and $|sc_1|$ is the distance from the radar to the reflector point.  

Applying the cosine law to the triangle formed by $S$, $C_1$, and $H$, we obtain:  

\begin{equation}
    |sh|^2 + |sc_1|^2 - 2 |sh| |sc_1| \cos(\theta^s_2-\theta^s_1) = |hc_1|^2.
\end{equation}  

\noindent where $\theta^s_2-\theta^s_1$ is the angle between $\Vec{sh}$ and $\Vec{sg_1}$
From the above relationships, we derive $|sc_1|$ as:  

\begin{equation}
    |sc_1| = \frac{2|sg'_1|^2 - 2|sg'_1||sh|}{2|sg'_1| - |sh|\cos(\theta^s_2-\theta^s_1) - |sh|}
    \label{eqn:ref_obj}
\end{equation}

The coordinates of the reflector point $C_1$ can then be determined as:  

\begin{equation}
    c_1 = s + \frac{\Vec{sg'}}{|sg'|} \cdot |sc_1|
\end{equation}  

where $\frac{\Vec{sg'}}{|sg'|}$ is the unit vector in the direction of $sg'$.  

\begin{figure}[t!]
    \centering
    \includegraphics[width=\linewidth, clip]{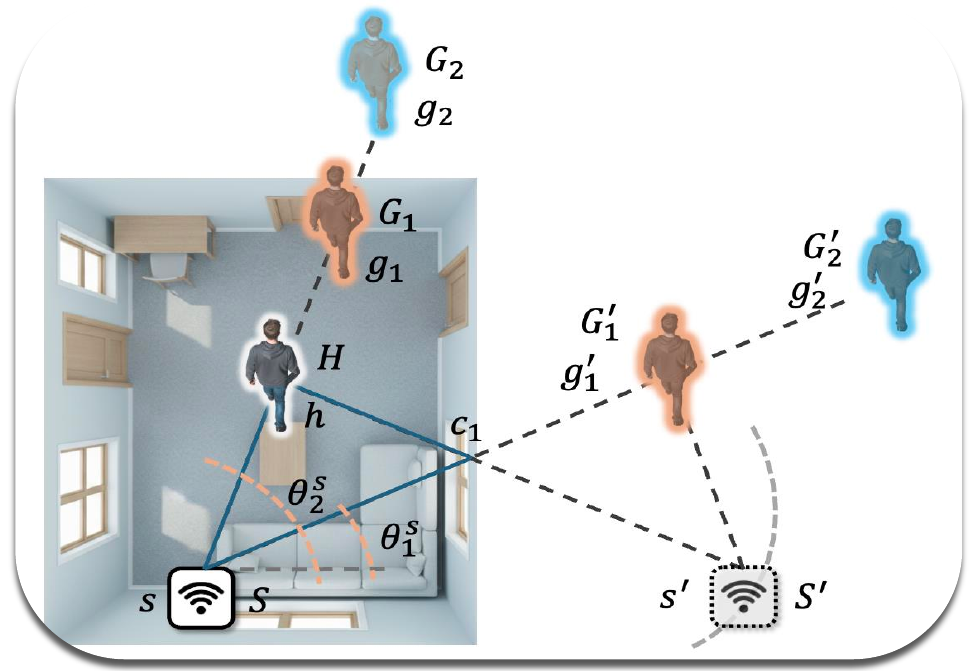}   
    \caption{\textbf{Multipath Inversion.}  
    Geometric relationships between ghost targets and reflectors. }
\label{fig:geometry}
\end{figure}

\subsubsection{Second-Bounce Ghost}

The second-bounce ghost point, denoted as \( G_2' \), follows a geometric relationship similar to the first-bounce case. The total path length from the radar to \( G_2' \) can be expressed as:

\begin{equation}
    |sg'_2| = |hc_1| + |sc_1|,
\end{equation}

where \( |sg'_2| \) represents the total distance from the radar to the second-bounce ghost point, \( |hc_1| \) is the distance from the human to the reflector, and \( |sc_1| \) is the distance from the radar to the reflector.

Using the cosine law and substituting the known relationships, we derive:

\begin{equation}
    |sc_1| = \frac{|sg'_2|^2 - |sh|^2}{2|sg'_1| - 2|sh|\cos(\theta_2^s - \theta_1^s)},
\end{equation}

\noindent where \( \theta_1^s \) and \( \theta_2^s \) denote the respective angles between the reflection points and the radar.

Subsequently, the remaining steps outlined in the first-bounce ghost section can be applied to determine the location of the mirror radar.

\subsection{Post Processing Reflector Points}
After estimating reflector points from ghost targets, we obtain a sparse and noisy set of 2D samples that approximate the underlying wall and object boundaries. To extract coherent geometric structures, we refine these points in two stages. 

First, we apply a Gaussian Mixture Model (GMM) to group the reflector points into spatially coherent clusters. The GMM models the distribution of reflector points as a mixture of Gaussian components, which enables soft probabilistic assignments and naturally captures elongated point distributions associated with physical surfaces. This clustering step separates points belonging to different walls or objects while remaining robust to uneven sampling and local noise.

Next, for each cluster, we perform RANSAC line fitting to recover the dominant structural direction and reject outliers introduced by multipath noise or incomplete reflections. RANSAC iteratively samples minimal point sets, fits line hypotheses, and selects the model with the largest inlier set under a point-to-line distance threshold. This procedure produces clean and geometrically consistent line segments for each cluster.

The resulting set of RANSAC-refined line structures forms a robust initial layout estimate, which is subsequently fed into our diffusion-based refinement module to obtain complete wall boundaries and object footprints.

\section{Discussion}

RISE demonstrates that a single static mmWave radar can reliably recover indoor layout and object information by leveraging multipath reflections and generative modeling. While our system has several limitations, we emphasize that it remains highly practical and broadly useful in many real-world settings.

A key limitation is that RISE relies on human motion to stimulate diverse multipath paths. This motion provides the geometric variation necessary for uncovering occluded regions, and completely static environments remain challenging. However, this requirement aligns well with many deployment scenarios: homes, clinics, elder-care facilities, offices, and rehabilitation environments naturally contain frequent human movement. In such settings, RISE can operate fully passively and unobtrusively, without requiring additional devices, active user participation, or intrusive sensors. Even short or routine movements, walking through a hallway or moving to sit in a chair, can generate sufficient multipath diversity for reliable reconstruction.

Second, blind regions inherent to fixed sensor placement cannot be eliminated entirely, although our bi-angular enhancement significantly reduces them. These blind spots typically correspond to uncommon reflection geometries and often do not affect the global layout.

Third, the current system outputs 2D top-down geometry rather than full 3D meshes or semantic object models. While this limits fine-grained reconstruction, top-down geometry is already sufficient for many high-impact applications such as indoor navigation, elder monitoring, health assessment, fall detection, and safety analysis—domains where privacy constraints prohibit cameras.

Despite these limitations, RISE remains an effective, privacy-preserving perception system capable of recovering meaningful indoor structure from a single static radar. It offers a lightweight, low-cost alternative to camera- or LiDAR-based approaches, especially in privacy-sensitive environments. These strengths, combined with its ability to operate passively using natural human motion, underscore its practical value and establish a foundation for future advances in radar-based indoor understanding.
\newpage
\newpage

\end{document}